\documentclass{llncs}

\usepackage{latexsym}
\usepackage{psfig}
\usepackage{url}

\newcommand{\datan}{\mathrm{DATALOG}^{\neg}}

\newcommand{\cc}{\mathit{cc}}
\newcommand{\sol}{\mathit{sol}}
\newcommand{\vp}{\mathit{vpos}}
\newcommand{\size}{\mathit{size}}
\newcommand{\pos}{\mathit{pos}}

\newcommand{\pr}{\mathit{Pr}}
\newcommand{\gr}{\mathit{gr}}

\newcommand{\at}{\mathit{At}}
\newcommand{\plc}{\mathit{PS}^\mathit{cwa}}
\newcommand{\eplc}{\mathit{PS}^+}

\newcommand{\vc}{\mathit{vc}}
\newcommand{\nq}{\mathit{nq}}
\newcommand{\invc}{\mathit{invc}}

\newcommand{\vtx}{\mathit{vtx}}
\newcommand{\edge}{\mathit{edge}}
\newcommand{\und}{{\_}}
\newcommand{\tr}{{\bf T}}
\newcommand{\fa}{{\bf F}}

\newcommand{\Ra}{\Rightarrow}

\newcommand{\lla}{\leftarrow}

\newcommand{\n}{\mathbf{not}}

\title{Propositional satisfiability in answer-set programming}

\author{Deborah East \and Miros\l aw Truszczy\'nski}

\institute{Department of Computer Science, University of Kentucky\\
           Lexington KY 40506-0046, USA\\
\email{\{deast, mirek\}@cs.engr.uky.edu}}

\begin{document}

\maketitle

\begin{abstract}
We show that propositional logic and its extensions can support
answer-set programming in the same way stable logic programming and
disjunctive logic programming do. To this end, we introduce a logic 
based on the logic of propositional schemata and on a version of the
Closed World Assumption. We call it the {\em extended logic of 
propositional schemata with CWA} ($\eplc$, in symbols). An important 
feature of the logic $\eplc$ is that it supports explicit modeling 
of constraints on cardinalities of sets. In the paper, we characterize 
the class of 
problems that can be solved by finite $\eplc$ theories. We implement 
a programming system based on the logic $\eplc$ and design and implement 
a solver for processing theories in $\eplc$. We present encouraging 
performance results for our approach --- we show it to be competitive 
with {\em smodels}, a state-of-the-art answer-set programming system 
based on stable logic programming. 
\end{abstract}

\section{Introduction}

Logic is most commonly used in declarative programming and knowledge
representation as follows. To solve a problem we represent its 
constraints and the relevant background knowledge as a theory in
the language of some logic. We formulate the goal (the statement of
the problem) as a formula of the logic. We then use proof techniques to
decide whether this formula follows from the theory. A proof of
the formula, variable substitutions or both determine a solution.

Recently, an alternative way in which logic can be used in computational 
knowledge representation has emerged from studies of nonstandard 
variants of logic programming such as logic programming with negation 
and disjunctive logic programming \cite{mt99,nie99}. This alternative 
approach is rooted in semantic notions and is based on methods to compute 
models. To represent a problem, we design a finite theory so that its {\em 
models} (and not proofs or variable substitutions) determine problem 
solutions (answers). To solve the problem, we compute models of the 
corresponding theory\footnote{We commonly restrict the language by 
disallowing function symbols to guarantee finiteness of models of finite 
theories. In the present paper, we also adopt this assumption.}. This 
model-based approach is now often referred to as {\em answer-set 
programming} (or ASP).

Logic programming with stable model semantics \cite{gl88} ({\em stable
logic programming} or {\em SLP}, in short) is an example of an ASP
formalism \cite{mt99}. In SLP, we represent problem constraints by a 
fixed program (independent of problem instances). We represent a specific 
instance of the problem (input data) by a collection of ground atoms. 
To solve the problem, we find stable models of the program formed jointly 
by the two components. To this end, we first {\em ground} it (compute 
its equivalent propositional representation) and, then, compute stable 
models of this grounded propositional program. Thanks to the emergence of 
fast systems to compute stable models of propositional logic programs, 
such as {\em smodels} \cite{ns00}, SLP is quickly becoming a viable 
declarative programming environment for computational knowledge 
representation. Disjunctive logic programming with the semantics of
answer sets \cite{gl90b} is another logic programming formalism that
fits well into the answer-set programming paradigm. An effective
solver for computing answer sets of disjunctive programs, {\em dlv},
is available \cite{elmps98} and its performance is comparable with that
of {\em smodels}. 

Our goal in this paper is to propose answer-set programming formalisms
based on propositional logic and its extensions. Our approach is
motivated by recent improvements in the performance of satisfiability
checkers. Researchers developed several new and fast implementations of 
the basic Davis-Putnam method such as {\em satz} \cite{la97} and {\em 
relsat} \cite{bs97}. A renewed interest in local-search techniques 
resulted in highly effective (albeit incomplete) satisfiability checkers 
such as {\em WALKSAT} \cite{skc94}, capable of handling large CNF 
theories, consisting of millions of clauses. Improvements in the 
performance resulted in an expanding range of applications of 
satisfiability checkers, with planning being one of the most 
spectacular examples \cite{kms96,ks99}. 

The way in which propositional satisfiability solvers are used in 
planning \cite{kms96} clearly fits the ASP paradigm. Planning problems
are encoded as propositional theories so that models correspond to
plans. In our paper, we extend ideas proposed in \cite{kms96} in the 
domain of planning and show that propositional satisfiability can be 
used as the foundation of a general purpose ASP system. To this end, we 
propose a logic to serve as a modeling language. This logic is a 
modification of the logic of propositional schemata \cite{kms96}; we 
explicitly separate theories into data and program, and use a version 
of Closed World Assumption (CWA) to define the semantics. This logic 
is nonmonotonic. We call it the {\em logic of propositional schemata 
with CWA} (or, $\plc$). 

The logic $\plc$ offers only basic logical connectives to help model 
problem constraints. We extend logic $\plc$ to support direct 
representation of constraints involving cardinalities. Examples of such
constraints are: "at least $k$ elements from the list must be in the
model" or "exactly $k$ elements from the list must be in the model".
They appear commonly in statements of constraint satisfaction problems.
We refer to this new logic as {\em extended logic of propositional 
schemata with Closed World Assumption} and denote it by $\eplc$. 

In the paper we characterize the class of problems that can be solved
by {\em finite} $\eplc$ theories. In other words, we determine the 
expressive power of the logic $\eplc$. Specifically, we show that it is 
equal to the expressive power of function-free logic programming with 
the stable-model semantics. 

For processing, theories in $\eplc$ could be compiled into propositional 
theories and ``off-the-shelf'' satisfiability checkers could be used
for processing. However, propositional representations of constraints 
involving cardinalities are usually very large and the sizes of the 
compiled theories limit the effectiveness of satisfiability checkers, 
even the most advanced ones, as processing engines. Thus, we argue against 
the compilation of the cardinality constraints. Instead, we propose an 
alternative approach. We design a ``target'' propositional logic for the 
logic $\eplc$ (propositional logic $\eplc$). In this logic, cardinality 
constraints have explicit representations and, therefore, do not need to 
be compiled any further. We develop a satisfiability checker for the
propositional logic $\eplc$ and use it as the processing 
back-end for the logic $\eplc$. Our solver is designed along the same 
lines as most satisfiability solvers implementing the Davis-Putnam 
algorithm but it takes a direct advantage of the cardinality
constraints explicitly present in the language. 

Experimental results on the performance of the overall system are 
highly encouraging. We obtain concise encodings of constraint problems
and the performance of our solver is competitive with the performance
of {\em smodels} and of state-of-the-art complete satisfiability 
checkers. Our work demonstrates that building propositional solvers
capable of processing of high-level constraints is a promising research 
direction for the area of propositional satisfiability.

Our paper is organized as follows. In the next section we introduce the
logic $\plc$ --- a fragment of the logic $\eplc$ without cardinality 
constraints. We determine the expressive power of the logic $\plc$ in
Section \ref{eps}. We discuss the full logic $\eplc$ in Section
\ref{eplc}. In the subsequent section we discuss implementation details 
and experimental results. The last section of the paper contains
conclusions and comments on the future work.

\section{Basic logic $\plc$}
\label{plc}

Our approach is based on the logic of {\em propositional schemata}.
The syntax of this logic is that of first-order logic without function 
symbols. The semantics is that of {\em Herbrand interpretations} and 
{\em models}, which we identify with subsets of the {\em Herbrand base}.
In the paper we consider only those theories in which at least one
constant symbol appears. Among all formulas in the language, of main 
interest to us are {\em clauses}, that is, expressions of the form
\begin{eqnarray}\label{cl-1}
a_1 \wedge \ldots \wedge a_m \Rightarrow
B_1 \vee \ldots \vee B_n,
\end{eqnarray}
where each $a_i$ is an atom and each $B_j$ is an atom or an 
expression of the form $\exists Y b(s)$, where $b(s)$ is an atom and 
$Y$ is a tuple 
of (not necessarily all) variables appearing in $b(s)$. Each of $m$ and 
$n$ (or both) may equal 0. If $m=0$, we replace the conjunct in the 
antecedent of the clause with a special symbol $\tr$ ({\em truth}). 
If $n=0$, we replace the empty disjunct in the consequent of the clause 
with a special symbol $\fa$ ({\em contradiction}). We assume that each 
clause is universally quantified and drop the universal
quantifiers from the notation. We further simplify the notation
by replacing each expression $\exists Y b(s)$ in the 
antecedent by $b(s')$, where in $s'$ we write a special 
symbol `$\und$' for each variable from $Y$ in $s$.

Let $T$ be a {\em finite} theory consisting of clauses. For a formula 
$B = \exists Y b(s)$ appearing in the consequent of a clause in $T$, 
we define $B^e$ to be the disjunction $B^e=\ \ \ b(s^1)\vee\ldots\vee 
b(s^k)$,
where $s^i$, $1\leq i\leq k$, range over all term tuples that can be 
obtained from $s$ by replacing variables in $Y$ with constants appearing
in  $T$. Since $T$ is finite, the disjunction is well defined (it has 
only finitely many disjuncts). 

For a clause $C \in T$ of the form 
(\ref{cl-1}), we define a clause $C^e$ by
\begin{eqnarray}\label{cl-2}
C^e=\ \ \ a_1 \wedge \ldots \wedge a_m \Rightarrow
B_1^e \vee \ldots \vee B_n^e,
\end{eqnarray}
A {\em ground instance} of $C$ is any formula obtained from 
$C^e$ by replacing every variable in $C^e$ by a constant appearing in
$T$ (different occurrences of the same variable must be replaced by 
the same constant).
We define the {\em grounding} of $T$, $\gr(T)$ as the collection of all 
ground instances of clauses in $T$, except for tautologies; they are
not included in $\gr(T)$. We have the following well-known result.
\begin{proposition}
Let $T$ be a finite clausal theory. Then a set of ground atoms $M$ is a
Herbrand model of $T$ if and only if $M$ is a (propositional) model of
$\gr(T)$.
\end{proposition}

The language may contain several {\em predefined} predicates and function 
symbols such as the equality operator and arithmetic comparators and 
operations. We assign to these symbols their standard interpretation.
However, we emphasize that the domains are restricted only to those 
constants that appear in a theory. 

We evaluate all expressions involving predefined function symbols and all 
atoms involving predefined relation symbols in the grounding process. If 
any argument of a predefined relation is not of the appropriate type, we 
interpret the corresponding atom as false. If a function yields as a 
result a constant that does not appear in the theory or if one of its 
arguments is not of the required type, we also interpret the corresponding 
atom as false. We then eliminate tautologies and simplify
the remaining clauses by removing true ``predefined'' atoms from 
the antecedents and false ``predefined'' atoms from the consequents. 
 
Let us consider an example. Let $T$ be a theory consisting of 
the following two clauses:
\begin{quote} 
$C_1 =\ \ q(b,c) \Ra p(a)$\\
$C_2 =\ \ p(X)\Ra(\exists Y q(X,Y)) \vee (X=a)$.
\end{quote}
There are three constants, $a$, $b$ and $c$, and two predicate symbols, 
$p$ and $q$, in the language. Symbols $X$ and $Y$ denote variables.
The clause $C_2$ can also be written (using the simplified notation) as
\begin{quote}
$C_2 =\ \ p(X)\Ra q(X,\und) \vee (X=a)$.
\end{quote}

To compute $\gr(T)$ we need to compute all ground instances of $C_2$
($C_1$ is itself its only ground instance). First, we compute 
the formula $C_2^e$:
\[
C_2^e = \ \ p(X)\Ra q(X,a)\vee q(X,b)\vee q(X,c)\vee (X=a).
\]
To obtain all ground instances of $C_2$ (or $C_2^e$), we replace $X$
with $a$, $b$ and $c$. The first substitution results in a
tautology (due to occurrence of `$a=a$' in the consequent of the
clause). Two other substitutions yield the following two ground
instances of $C$ (we drop atoms `$b=a$' and `$c=a$' from the 
consequents; they are false by the standard interpretation of 
equality):
\begin{quote}
$p(b)\Ra q(b,a)\vee q(b,b)\vee q(b,c)$\\
$p(c)\Ra q(c,a)\vee q(c,b)\vee q(c,c)$.
\end{quote}
These two clauses together with $C_1$ form $\gr(T)$. The sets of ground 
atoms $\{p(a), q(b,c)\}$ and $\{p(b),p(c),q(b,a),q(c,c)\}$ are two 
examples of models of $T$ (or $\gr(T)$). 

In order for the logic of propositional schemata to be useful as a 
programming tool, we modify it to separate input data from the 
program encoding the problem to be solved. We distinguish in the set 
of predicates $\pr$ of the language a subset, $\pr'$. We call its 
elements {\em data} predicates. We assume that predefined predicates are
{\em not} data predicates. All predicates other than data predicates and 
predefined predicates are called {\em program} predicates. A {\em theory} 
of our logic is a pair $(D,P)$, where $D$ is a {\em finite} collection 
of ground atoms whose predicate symbols are data predicates ({\em data}), 
and $P$ is a {\em finite} collection of clauses (a {\em program}). 

To define the semantics for the logic, we use grounding and a
form of CWA. We say that a set of ground atoms (built of 
data and program predicates) is a {\em model} of a theory $(D,P)$ if 
\newcounter{ct8}
\begin{list}{M\arabic{ct8}:\ }{\usecounter{ct8}\topsep 0.03in
\parsep 0in\itemsep 0in}
\item $M$ is a model of $\gr(D\cup P)$ (or, equivalently, $M$ is
an Herbrand model of $D\cup P$), and
\item for every ground atom $p(t)$ such that $p\in \pr'$ ($p$ is 
a data predicate), $p(t)\in M$ if and only if $p(t)\in D$.
\end{list}

We call the logic described above the {\em logic of propositional schemata 
with CWA} and denote it by $\plc$. Due to (M2), not every model
of $\gr(D,P)$ is a model of $(D,P)$. Consequently, one can show that our 
logic is nonmonotonic. This difference between the logic of propositional 
schemata and the logic $\plc$, while seemingly small, has significant 
consequences for the expressive power of the logic and its applicability 
as a programming tool. 

Before addressing these two issues, let us consider an example.
Let $A$ and $B$ be two disjoint and finite sets. We define
$D=\{p_1(a)\colon a\in A\}\cup \{p_2(b)\colon b\in B\}$. We define $P$
to consist of two clauses:\\
Ex1: $q_1(X) \Ra p_1(X)$\ \ \ \ \ \ Ex2: $q_2(X) \Ra p_2(X)$.


The constants are elements of $A\cup B$; $X$ is a variable.
The predicates are $p_1$, $p_2$, $q_1$ and $q_2$. The first two 
are data predicates.

By (M2), each model of a $\plc$ theory $(D,P)$ contains $D$. However, 
it does not contain any ground atom $p_1(b)$, where $b\in B$, nor any 
ground atom $p_2(a)$, where $a\in A$. Each ground instance of the clause
(Ex1) is of the form $q_1(c)\Ra p_1(c)$, where $c$ is a constant ($c\in
A \cup B$). Since $p_1(c)\in M$ if and only if $c\in A$, it follows that
if $q_1(c)\in M$, then $c\in A$. Similarly, we obtain that if $q_1(c)\in
M$, then $c\in A$. Thus, $M$ is a model of $(D,P)$ if and only if
$M=D\cup\{q_1(a)\colon a\in A'\}\cup \{q_2(b)\colon b\in B'\}$, for some
$A'\subseteq A$ and $B'\subseteq B$.

Let us choose an element from $A$, say $a_0$, and an element from $B$, 
say $b_0$. Let us then add to $P$ the clause

\noindent
Ex3: $p_1(a_0) \Ra p_1(b_0)$

\noindent
We denote the new program by $P'$. The $\plc$ theory $(D,P)$ has no
models even though $gr(D,P)$ is propositionally consistent. The reason is
that all propositional models satisfying $\gr(D,P)$ contain $p_1(b_0)$.
Thus, none of these models satisfies condition (M2). This example
illustrates that our semantics is different from circumscription as
circumscription preserves consistency. Circumscription applied to $p_1$
would result in models in which the extension of $p_1$ in $D$ would
be minimally extended by one more constant $b_0$. Our (strong)
minimization principle does not allow for any additions to the extension
of data predicates. Intuitively, it is exactly as it should be. {\bf Data
predicates are meant to represent input data. The program should not be 
able to extend it.}

Logic $\plc$ is a tool to model problems. To illustrate this use
of the logic, we show how to encode the {\em vertex-cover} problem for
graphs. Let $G=(V,E)$ be a graph. A set $W\subseteq V$ is a {\em 
vertex cover} of $G$ if for every edge $\{x,y\}\in E$, $x$ or $y$ (or 
both) are in $W$. The vertex-cover problem is defined as follows: given 
a graph $G=(V,E)$ and an integer $k$, decide whether $G$ has a vertex 
cover with no more than $k$ vertices. 

For the vertex-cover problem the input data is described by the
following set of ground atoms: 

\noindent 
$D_\vc = \{\vtx(v)\colon v\in V\} \cup \{\edge(v,w)\colon \{v,w\}\in E\}
       \cup \{\size(k)\} \cup \{\pos(i)\colon 1,\ldots, n\}$.

\noindent
This set specifies the set of vertices and the set of edges of an input
graph. It provides the limit on the size of a vertex cover sought.
Lastly, it uses a predicate $\pos$ to specify a range of {\em integers} 
that will be used to label vertices. The problem itself is described by 
the program $P_\vc$:

\newcounter{ct2}
\begin{list}{VC\arabic{ct2}:\ }{\usecounter{ct2}\topsep 0.03in
\parsep 0in\itemsep 0in}
\item $\vp(I,X) \Rightarrow \vtx(X)$
\item $\vp(I,X)\Rightarrow \pos(I)$
\item $\vtx(X)\Rightarrow \vp(\und,X)$
\item $\vp(I,X)\wedge\vp(J,X) \Rightarrow I=J$
\item $\vp(I,X)\wedge\vp(I,Y) \Rightarrow X=Y$
\item $\edge(X,Y) \wedge \vp(I,X) \wedge \vp(J,Y) \wedge \size(K) 
\Rightarrow (I\leq K) \vee (J\leq K)$
\end{list}

(VC1) and (VC2) ensure that $\vp(i,x)$ is false if $i$ is 
not an integer from the set $\{1,\ldots, n\}$ or if $x$ is not a vertex. 
(VC3)-(VC5) together enforce that the atoms $\vp(i,x)$ that 
are true in a model of the $\plc$ theory $(D_\vc,P_\vc)$ define a 
permutation of the vertices in $V$. Finally, (VC6) ensures that 
each edge has at least one vertex assigned by $\vp$ to positions 
$1,\ldots,k$ (in other words, that vertices labeled $1,\ldots, k$ form 
a vertex cover). The correctness of this encoding is formally
established in the following result. 

\begin{proposition}
\label{p-vtx}
Let $G=(V,E)$ be an undirected graph and let $k$ be a positive integer.
A set of vertices $\{w_1,\ldots,w_{k}\}\subseteq V$ is a vertex cover of 
$G$ if and only if $M = D_\vc \cup \{\vp(i,w_i) \colon i=1,\ldots,k\}$ 
is a model of the theory $(D_\vc,P_\vc)$. 
\end{proposition}

For another example, we will consider the $n$-queens problem, that is,
the problem of placing $n$ queens on a $n\times n$ chess board so that
no queen attacks another.

In this case, the representation of input data describes the set of row
and column indices:

\noindent
$D_\nq = \{\pos(i)\colon 1,\ldots, n\}$.

\noindent
The problem itself is described by the program $P_\nq$. The predicate
$q$ describes a distribution of queens on the board: $q(x,y)$ is true
precisely when there is a queen in the position $(x,y)$.

\newcounter{ct17}
\begin{list}{nQ\arabic{ct17}:\ }{\usecounter{ct17}\topsep 0.03in
\parsep 0in\itemsep 0in}
\item $q(R,C) \Rightarrow \pos(R)$
\item $q(R,C) \Rightarrow \pos(C)$
\item $q(R,C1)\wedge q(R,C2) \Rightarrow C1=C2$
\item $q(R1,C)\wedge q(R2,C) \Rightarrow R1=R2$
\item $q(R,C), q(R+I,C+I) \Rightarrow {\bf F}$
\item $q(R,C), q(R+I,C-I) \Rightarrow {\bf F}$
\end{list}

The first two clauses ensure that if $q(r,c)$ is true in a model of 
$(D_\nq,P_\nq)$ then $r$ and $c$ are integers from the set
$\{1,\ldots,n\}$. The following two clauses enforce the constraint that
no two queens are placed in the same row or the same column. Finally,
the last two clauses guarantee that no two queens are placed on the
same diagonal. As in the case of the vertex cover problem, also in this
case we can formally show the correctness of this encoding. 

These examples demonstrate that $\plc$ programs can serve as
representations of computational problems. Two key questions arise:
(1) what is the expressive power of the logic $\plc$, and (2) how to
use the logic $\plc$ as a practical computational tool. We address 
both questions in the remainder of the paper.

\section{Expressive power of $\plc$}
\label{eps}

A {\em search} problem, $\Pi$, is given by a set of finite {\em 
instances}, $D_\Pi$, such that for each instance $I \in D_\Pi$, there 
is a finite set $S_\Pi(I)$ of all {\em solutions} to $\Pi$ for the 
instance $I$ \cite{gj79}. The graph-coloring, vertex-cover and
$n$-queens problems considered in the previous section are search 
problems. More generally, 
all constraint satisfaction problems including basic AI problems such 
as planning, scheduling and product configuration can be cast as search 
problems. 

We say that a $\plc$ program $P$ {\em solves} a search problem $\Pi$ 
if there exist:

\newcounter{ct4}
\begin{list}{\arabic{ct4}.}{\usecounter{ct4}\topsep 0.03in
\labelwidth 0.1in\itemindent 0in\labelsep 0.1in\leftmargin 0.2in
\parsep 0in\itemsep 0in}
\item A mapping $d$ that can be computed in polynomial time and that
encodes instances to $\Pi$ as sets of ground atoms built of data 
predicates
\item A partial mapping $\sol$, computable in polynomial time, 
that assigns to (some) sets of ground atoms solutions 
to $\Pi$ (elements of $\bigcup_{I\in D_{\Pi}} S_\Pi(I)$) 
\end{list} 
such that for every instance $I\in D_{\Pi}$, $s\in S_\Pi(I)$ if and only 
if there exists a model $M$ of the $\plc$ theory $(d(I),P)$ such that 
$M$ is in the domain of the mapping $\sol$ and $\sol(M) =s$.


A search problem $\Pi$ is in the class {\em NP-search} if there is 
a nondeterministic Turing Machine $\mathit{TM}$ such that (1)
$\mathit{TM}$ runs in polynomial time; (2)
for every instance $I\in D_\Pi$, the set of strings left on the
tape when accepting computations for $I$ terminate is precisely the set
of solutions $S_\Pi(I)$.

We now have the following theorem that determines the expressive power
of the logic $\plc$. Its proof is provided in the appendix.

\begin{theorem}
\label{ep-main}
A search problem $\Pi$ can be solved by a $\plc$ program if and only if
$\Pi\in$ NP-search.
\end{theorem}

Decision problems can be viewed as special search problems. For the class 
of decision problems, Theorem \ref{ep-main} implies the following corollary
(a counterpart to the result on the expressive power of $\datan$
\cite{sch91}).

\begin{corollary}
A decision problem $\Pi$ can be solved by a $\plc$ program if and only
if $\Pi$ is in NP.
\end{corollary}

\section{Extending $\plc$ --- the logic $\eplc$}
\label{eplc}

We will now discuss ways to enhance effectiveness of logic $\plc$ as
a modeling formalism and propose ways to improve computational 
performance. When considering the $\plc$ theories developed for the
$n$-queens and vertex-cover problems one observes that these theories
could be simplified if the language of the logic $\plc$ contained direct
means to model constraints such as: ``exactly one element is selected'' or 
``at most $k$ elements are selected''.

With this motivation, we extend the language of the logic $\plc$. We 
define a {\em c-atom} (cardinality atom) as an expression
$m\{p(X,\_,Y)\}n$,
where $m$ and $n$ are non-negative integers, $X$ and $Y$ are tuples of 
variables and $p$ is a program predicate\footnote{In our
implementation, we support a somewhat more general form of c-atoms.}. 

The interpretation of a c-atom is that for every ground tuples $x$ and 
$y$ that can be substituted for $X$ and $Y$, at least $m$ and at most 
$n$ atoms from the set
\[
\{p(x,c,y)\colon c \ \mbox{is a constant appearing in the theory}\}
\]
are true. One of $m$ and $n$ may be missing from the expression. If $m$
is missing, there is no lower-bound constraint on the number of atoms 
that are true. If $m$ is missing, there is no upper-bound constraint on the
number of atoms that are true. It is also possible to have more
``underscore'' symbols in c-atoms. In such case, when forming the set
of atoms on which cardinality constraints are imposed, all possible
ways to replace the ``underscore'' symbols by constants are used.

An {\em extended clause} is a clause built of c-atoms. The notions of 
a {\em program} and {\em theory} are defined as in the case of the logic 
$\plc$.

A theory in the extended syntax can be grounded, that is, represented 
as a set of propositional clauses, in a similar way as before. In 
particular, data and predefined predicates are treated in the same way and 
are subject to the same version of CWA that was used for the
logic $\plc$. While grounding, c-atoms are interpreted as explained
earlier. Grounding allows us to lift the semantics of propositional logic 
to the theories in the extended syntax. We call the resulting logic the 
{\em extended logic $\plc$} and denote it by $\eplc$.

In the logic $\eplc$ we can encode the vertex cover problem in a more
straightforward and more concise way. Namely, the problem can be 
represented without the need for integers to label the vertices of an 
input graph! This new representation $(D'_\vc,P'_\vc)$ is given by: 

\noindent
$D'_\vc = \{\vtx(v)\colon v\in V\} \cup \{\edge(v,w)\colon \{v,w\}\in E\}$,

\noindent
and $P'_\vc =$

\setcounter{ct2}{0}
\begin{list}{VC$'$\arabic{ct2}:\ }{\usecounter{ct2}\topsep 0.03in
\parsep 0in\itemsep 0in}
\item $\invc(X)\Rightarrow \vtx(X)$
\item $\{\invc(\und)\}k$ 
\item $\edge(X,Y) \Rightarrow \invc(X)\vee \invc(Y)$. 
\end{list}

Atoms $\invc(x)$ that are true in a model of the $\plc$ theory $(D'_\vc,
P'_\vc)$ define a set of vertices that is a candidate for a vertex cover.
(VC$'$2) guarantees that no more than $k$ vertices are included. 
(VC$'$3) enforces the vertex-cover constraint.

We close this section with an observation on the expressive power of
the logic $\eplc$. Since it is a generalization of the logic $\plc$,
it can capture all problems that are in the class NP-search. On the
other hand, the problem of computing models of a $\eplc$ theory with a 
fixed program part is an NP-search problem, it follows that the 
expressive power of the logics $\eplc$ does not extend beyond the class
NP-search. In other words, the logic $\eplc$ also captures the class 
NP-search. 

\section{Computing with $\eplc$ theories}

To process $\eplc$ theories, one approach is to ground them into 
collections of propositional clauses. However, CNF representations of 
c-atoms may be quite large; the constraint ``at most $n$ atoms in the 
set $\{p_1,\ldots, p_k\}$ are true'', is captured by $\theta(k^{n+1})$ 
clauses $p_{i_1},\ldots,p_{i_{n+1}} \Rightarrow \fa$,
one for each $(n+1)$-element subset $\{p_{i_1},\ldots,p_{i_{n+1}}\}$ 
of $\{p_1,\ldots,p_k\}$.
 
Thus, we propose another approach. The idea is to 
develop an extension of propositional logic representing 
c-atoms directly. Let $\at$ be a set of propositional variables. By 
a {\em propositional c-atom} we mean any expression of the 
form
$m\{p_1,\ldots,p_k\}n$,
where $m$ and $n$ are non-negative integers and $p_1,\ldots,p_k$ are 
atoms in $\at$ (one of $m$ and $n$ may be missing). By an {\em extended
propositional clause} we mean an expression of the form
\[
C=\ \ A_1\wedge\ldots \wedge A_s\Rightarrow B_1\vee\ldots\vee B_t,
\]
where all $A_i$ and $B_i$ are propositional c-atoms.

Let $M\subseteq \at$ be a set of atoms. We say that $M$ {\em satisfies}
a generalized atom $m\{p_1,\ldots,p_k\}n$ if 
\[
m \leq |M\cap \{p_1,\ldots,p_k\}|\leq n.
\]
Further, $M$ {\em satisfies} a generalized clause $C$ if $M$ satisfies
at least one atom $B_j$ or does not satisfy at least one atom $A_i$.
We call the resulting logic the {\em propositional logic} $\eplc$.
Clearly, $M$ satisfies an atom $1\{p\}1$ if and only if $p\in M$. Thus,
the propositional logic $\eplc$ extends the (clausal) propositional logic.

Theories of the logic $\eplc$ can be grounded in the extended 
propositional logic by generalizing the approach described in Section
\ref{plc}. We represent c-atoms as propositional c-atoms and avoid a 
blow-up in the size of the representation. The problem is that SAT 
checkers cannot now be used to resolve the satisfiability of the extended 
propositional logic as they are not designed to work with the extended
syntax. 

It is clear, however, that the techniques developed in the area of 
SAT checkers can be extended to the propositional logic $\eplc$. We 
have developed a Davis-Putnam like procedure, {\em aspps}, that finds 
models of propositional $\eplc$. We also developed a program {\em psgrnd} 
that accepts theories in the syntax of the logic $\eplc$ and grounds them
into propositional $\eplc$ theories. Thus, the two programs together can 
be used as a processing mechanism for an answer-set programming system 
based on the logic $\eplc$. The programs {\em psgrnd} and {\em aspps} are 
available at \url{http://www.cs.uky.edu/ai/aspps/}.

In our experiments we considered the vertex-cover problem and several
combinatorial problems including $n$-queens problem, pigeonhole
problem and the problem to compute Schur numbers. All our experiments
were performed on a Pentium III 500MHz machine running linux.

We were mostly interested in comparing the performance of our system
{\em psgrnd/aspps} with that of {\em smodels}. The reason is that both
programs accept similar syntax and allow for very similar modeling of
constraints. We also experimented with a satisfiability checker
{\em satz}. 

In the case of vertex cover, for each $n=50$, 60, 70 and 80, we randomly 
generated 100 graphs with $n$ vertices and $2n$ edges. For each graph 
$G$, we computed the minimum size $k_G$ for which the vertex cover can 
be found. We then tested {\em aspps}, {\em smodels} and {\em satz} on all 
the 
instances $(G,k_G)$. The results represent the average execution times 
Encodings we used for testing {\em aspps} and {\em smodels} where based on 
the clauses (VC$'$1) - (VC$'$3). For {\em satz} we used encodings based on
the clauses (VC1) - (VC6) (cardinality constraints cannot be handled
by {\em satz}).

A propositional CNF theory obtained by grounding the program (VC1) - 
(VC6), has $\Theta(n^2)$ atoms, $\Theta(mn^2)$ clauses and its total 
size is also $\Theta(mn^2)$. For input instances we used in our
experiments, these theories were of such large sizes (over one million
rules in the case of graphs with 80 vertices) that {\em satz} did not 
terminate in the time we allocated (5 minutes). Thus, no times for
{\em satz} are reported. On the other hand, since the propositional
$\eplc$ theory obtained by grounding the $\eplc$ program (VC$'$1) - 
(VC$'$3) has only 
$\Theta(m+n)$ clauses (a few hundred clauses for graphs with 80
vertices) and its total size has the same asymptotic estimate. This is 
dramatically less than in the case of theories {\em satz} had to
process. Both {\em aspps} and {\em smodels} performed very well, with {\em 
aspps} being about three times faster than {\em smodels}. The timing results 
are summarized in Table 1.

{\small
\begin{center}
\begin{tabular}{|l|r|r|r|r|}
\hline
$n$ & 50 & 60 & 70 & 80 \\
\hline
{\em aspps} & 0.04 & 0.22 & 1.26 & 6.45 \\
\hline
{\em smodels} & 0.12 & 0.76 & 4.14 & 22.35 \\
\hline
\end{tabular}
\ \\
\ \\
{\bf Table 1}. Timing results (in seconds) for the vertex-cover
problem.
\end{center}
}
     
For the $n$-queens problem, our solver performed exceptionally well. It 
scaled up much better than {\em smodels} both in the case when we were looking 
for one solution and when we wanted to compute all solutions. In
particular, our program found a solution to the 36 queens problem in
0.97 sec. It also outperformed {\em satz}. 

{\small
\begin{center}
\begin{tabular}{|l|r|r|r|r|r|r|}
\hline
\# of queens & 18 & 19 & 20 & 21 & 22 & 23 \\
\hline
{\em aspps} & 0.02 & 0.02 & 0.07 & 0.07 & 0.11 & 0.12 \\
\hline 
{\em smodels} & 2.35 & 1.28 & 13.25 & 19.31 & 167.1 & 380.35 \\
\hline 
{\em satz} & 1.16 & 0.61 & 4.35 & 0.95 & 28.64 & 1.42 \\
\hline
\end{tabular}
\ \\
\ \\
{\bf Table 2.} Timing results (in seconds) for the $n$-queen problem.
\end{center}
}

The pigeonhole problem consists of showing that it is not possible to
place $p$ pigeons in $h$ holes if $p>h$. For this problem {\em aspps}
showed the best performance --- about three times faster than the other
two solvers (all programs showed a similar rate of growth in the execution 
time).

{\small
\begin{center}
\begin{tabular}{|l|r|r|r|r|}
\hline
$(p,h)$ & (9,8) & (10,9) & (11,10) & (12,11) \\
\hline
{\em aspps} & 0.59 & 5.63 & 60.08 & 702.02  \\
\hline
{\em smodels} & 2.7 & 21.56 & 219.99 & 2469.97 \\
\hline
{\em satz} & 1.87 & 17.28 & 178.20 & 2044.42 \\
\hline
\end{tabular}
\ \\
\ \\
{\bf Table 3.} Timing results (in seconds) for the pigeonhole problem
\end{center}
}

The Schur problem consists of placing $n$ numbers $1,2,\ldots,n$ in 
$k$ bins so that the set of numbers assigned to a bin is not closed under 
sums. That is, for all numbers $x$, $y$, $z$, $1\leq x,y,z\leq n$, if 
$x$ and $y$ are in a bin $b$, then $z$ is not in $b$ ($x$ and $y$ need 
not be distinct). The Schur number $S(k)$ is the maximum number $n$ for 
which such a placement is still possible.

We considered the problem of the existence of the placement for $k=4$ 
and values of $n$ ranging from 40 to 45. For $n\leq 44$ all programs
found a ``Schur'' placement. However, no ``Schur'' placement exists for 
$n=45$ (and higher values of $n$). All programs were able to establish
the non-existence of solutions for $n=45$ (but the times grew
significantly). Our results summarizing the performance of our system and 
{\em smodels} on the theories encoding the constraints of the problem 
are shown in Table 4. {\em aspps} and {\em satz} seem to performed better
than {\em smodels}, with {\em satz} being slightly faster for values of
$n$ closer to the Schur number.

{\small
\begin{center}
\begin{tabular}{|l|r|r|r|r|r|r|}
\hline
$n$ & 40 & 41 & 42 & 43 & 44 & 45\\
\hline
{\em aspps} & 0.03 & 0.03 & 0.03 & 0.03 & 1.83 & 54.5 \\
\hline
{\em smodels} & 0.3 & 0.38 & 0.32 & 0.36 & 35.8 & $>$1500  \\
\hline
{\em satz} & 0.21 & 0.23 & 0.24 & 0.25 & 0.96 & 20.4\\
\hline
\end{tabular}
\ \\
\ \\
{\bf Table 4}. Timing results (in seconds) for the Schur-number problem.
\end{center}
}

In the case of the last three problems, it was possible to eliminate 
cardinality constraints without significant increase in the size of
grounded theories. As a result, {\em satz} performed well. 

\section{Conclusions}

Our work demonstrates that propositional logic and its extensions 
can support answer-set programming systems in a way in which stable 
logic programming and disjunctive logic programming do\footnote{We point 
out, though, that stable logic programming and disjunctive logic 
programming directly support negation-as-failure and, consequently,
yield more direct solutions to some knowledge representation problems
such as, for example, the frame problem.}.
In the paper we described logic $\eplc$ that can be used to this end.
We presented an effective implementation of a grounder, {\em psgrnd},
and a solver, {\em aspps}, for processing theories in the logic $\eplc$. 
Our experimental results are encouraging. Our system is competitive with 
{\em smodels}, and in many cases outperforms it. It is also 
competitive with satisfiability solvers such as {\em satz}. 

The results of the paper show that programming front-ends for 
constraint satisfaction problems that support explicit coding of complex 
constraints facilitate modeling and result in concise representations.
They also show that solvers such as {\em aspps} that take advantage of 
those concise encodings and process high-level constraints directly, 
without compiling them to simpler representations, exhibit very good 
computational performance. These two aspects are important. 
Satisfiability checkers often cannot effectively solve problems simply 
due to the fact that encodings they have to work with are large. For
instance, for the vertex-cover problem for graphs with 80 vertices and
160 edges, {\em aspps} has to deal with theories that consist of a few
hundred of rules only. In the same time pure propositional encodings 
of the same problem contain over one million clauses --- a factor that
undoubtedly is behind much poorer performance of {\em satz} on this 
problem.  
 
Our work raises new questions. Further extensions of logic $\eplc$ are 
possible. For instance, constraints that impose other conditions on 
set cardinalities than those considered here (such as, the {\em parity} 
constraint) might be included. We will pursue this direction. Similarly, 
there is much room for improvement in the area of solvers for the
propositional logic $\eplc$. In particular, we will study local search 
algorithms as possible satisfiability solvers for propositional $\eplc$ 
theories. 

Finally, we note that the experimental results presented here are 
meant to show that {\em aspps} is competitive with other solvers and, 
we think, they demonstrate this. However, these results are still 
too fragmentary to provide basis for any conclusive comparison between 
the three solvers tested. Such a comparison is further complicated 
by the fact that the same problem may have several different encodings 
with different computational properties.
Developing the methodology for comparing solvers designed to work with 
different formal systems is a challenging problem for builders of 
constraint solvers and declarative programming systems.



\section*{Acknowledgments}

This work was partially supported by the NSF grants CDA-9502645, 
IRI-9619233 and EPS-9874764.

{\small


\newcommand{\etalchar}[1]{$^{#1}$}

}

\section*{Appendix}

We will present here a sketch of a proof of our main result concerning 
the expressive power of the logic $\plc$. The proof relies on some basic
notions from logic programming (we refer the reader to
\cite{ap90,llo84} for details).

We restrict our discussion to function-free languages (the case
relevant to our logic $\plc$). Given a predicate language ${\cal L}$ 
(as defined in Section \ref{plc}), a {\em logic program clause} over this 
language is an expression $r$ of the form
\[
r =\ \ \ p(t) \lla q_1(t_1),\ldots, q_m(t_m),
           \n(q_{m+1}(t_{m+1})),\ldots, \n(q_{m+n}(t_{m+n}))
\]
where $p, q_1, \ldots, q_{m+n}\in \pr$, (we assume that $p$ is not a
predefined predicate), and $t,t_1,\ldots, t_{m+n}$ are term tuples 
with the arity matching the arity of the corresponding predicate 
symbol. We call the atom $p(t)$ the {\em head} of the rule $r$ and 
denote it by $h(r)$. For a rule $r$ we also define
\[
B(r) = q_1(t_1) \wedge \ldots q_m(t_m) \wedge \neg q_{m+1}(t_{m+1})
\wedge \ldots\wedge \neg q_{n}(t_{n})
\]

We will be interested in supported models of logic programs. Without
loss of generality, we will restrict our attention to programs in
the {\em normal form}. That is, we assume that (1) the head of each rule 
is of the form $p(t)$, where $t$ is a tuple of variables, and (2) if
$p$ appears in the head of two rules, the heads of these two rules are
exactly the same (the same tuple of variables appear in both of them)
\cite{cl78,ap90}. 

Let $P$ be a program in the normal form. For each predicate symbol 
$p\in \pr(P)$, we define a formula $\cc(p)$ by:
\[
\cc(p) =\ \ \ p(X)\Leftrightarrow \bigvee \{\exists Y_r B(r)
\colon r\in P', h(r)=p(X)\}, 
\]
where $X$ is a tuple of variables and $Y_r$ is the tuple of variables 
occurring in the body of $r$ but not in the head of $r$ (we exploit the
normal form of $P$ here). We define the {\em completion} of $P$, $CC(P)$, 
by setting $CC(P) = \{\cc(p)\colon p\in \pr\}$.

The Clark's completion is important as it allows us to characterize
{\em supported} models of a logic program \cite{ap90}. Namely, we have
the following result.

\begin{theorem}
\label{supm}
Let $P$ be a logic program. A set of ground atoms $M$ is a supported
model of $P$ if and only if it is a Herbrand model of $CC(P)$.
\end{theorem}

We now have the following theorem.

\begin{theorem}
\label{ep-aux}
Let $P$ be a logic program in the normal form. Let $\pr$ be the set of
predicates appearing in $P$ and let $\pr'$ be the set of predicates of
$P$ that do not appear in the heads of rules in $P$. There is a $\plc$
theory $T(P)$ such that for every set of ground atoms $D$ over
predicates from $\pr'$, a set of ground atoms $M$ is a supported model
of $D\cup P$ if and only if $M=M'\cap HB(P)$ for some model $M'$ of the 
$\plc$ theory $(D,T(P))$.
\end{theorem}
Proof: (Sketch) To define $T(P)$, we consider the completion $CC(P)$ of 
$P$. The idea is to take for $T(P)$ an equivalent clausal representation 
of $CC(P)$. 

We build such representation as follows. Let $p$ be a predicate symbol 
in $\pr\setminus\pr'$. The completion $CC(P)$ contains the formula 
\[
\cc(p) = \ \ p(X)\Leftrightarrow \bigvee\{\exists Y_r B(r)\colon 
r\in P, h(r)=p(X)\},
\]
where $X$ is a tuple of variables and $Y_r$ is the tuple of variables
occurring in the body of $r$ but not in the head of $r$. For each 
rule $r\in P$ such that $p$ occurs in $h(r)$, we introduce a new 
predicate symbol $d_{r}$, of the same arity $|X|+|Y_r|$. We define 
a theory $T'(P)$ to consist of the following formulas (we recall that
$B(r)$ stands for the conjunction of the literals from the body of $r$):
\[
\psi(r) = \ \ d_{r}(X,Y_r) \Leftrightarrow B(r),
\]
where $p\in \pr\setminus\pr'$, $r\in P$ and $p$ occurs in the 
head of $r$, and
\[
\cc'(p) = \ \ p(X)\Leftrightarrow \bigvee\{\exists Y_r d_{r}(X,Y_r) 
\colon r\in P, h(r)=p(X)\},
\]
where $p\in \pr\setminus \pr'$.

It is clear that the theory $T'(P)$ is equivalent to $CC(P)$ (modulo 
new ground atoms). That is, $M\subseteq HB(P)$ is a model of $CC(P)$
if and only if $M=M'\cap HB(P)$, for some model $M'$ of $T'(P)$. 

%

One can show that $T'(P)$ can be rewritten (in polynomial time)
into an equivalent clausal
form, $T(P)$. Consequently, $T(P)$ is equivalent to $CC(P)$ (modulo 
ground atoms $d_{r}(t)$). It is now a routine task to verify that the 
theory $T(P)$ satisfies all the requirements of the statement of the 
theorem. \hfill $\Box$

Using the terminology introduced here we will now prove Theorem
\ref{ep-main} from Section \ref{eps}.

\begin{theorem}
A search problem $\Pi$ can be solved by a finite $\plc$ program if and 
only if $\Pi\in$ NP-search.
\end{theorem}
Proof: (Sketch) In \cite{mr00} it is proved that every NP-search problem 
can be solved uniformly by a finite logic program under the supported-model 
semantics. Since the theory $T(P)$ can be constructed in polynomial time, it
follows by Theorem \ref{ep-aux} that every search problem in NP-search
can be solved by a finite $\plc$ program. Conversely, for every fixed 
program $P$, the problem of computing models of a $\plc$ theory $(D,P)$ 
($D$ is the input) is clearly in the class NP-search. Thus, only search
problem in the class NP-search can be solved by finite $\plc$ programs.
Hence, the assertion follows. \hfill$\Box$
\end{document}